\documentclass[11pt]{article}

\usepackage[margin=1in]{geometry}
\usepackage{amsmath,amssymb}
\usepackage{booktabs}
\usepackage{graphicx}
\usepackage{array}
\usepackage{caption}
\usepackage{microtype}
\usepackage[colorlinks=true,linkcolor=blue,citecolor=blue,urlcolor=blue]{hyperref}
\graphicspath{{figures/}}

\title{\bfseries What Does Context Compression Cost an Agent?\\[2pt]
       \large Interaction Costs Unrevealed by Task-Completion Metrics}

\author{%
  Shuyu Liu\\
  \normalsize Chongqing University of Science and Technology%
}
\date{2026}

\begin{document}

\maketitle

\begin{center}
\footnotesize
\textit{An earlier version of this work was accepted at the COLM 2026 Workshop on Efficient Reasoning.}
\end{center}

\begin{abstract}
\noindent
Task completion is the standard metric for evaluating context compression, yet it is an
incomplete measure: compression can substantially increase an agent's interaction cost ---
the reacquisition of state it dropped --- while leaving completion statistically unchanged.
We formalize completion-only evaluation as a lossy projection of the full evaluation
outcome, and demonstrate in a controlled setting that it can miss a large interaction
cost.

We introduce a controlled runtime measurement protocol that isolates the reacquisition
cost of context compression in a bounded-horizon agent. An agent acts on a deterministic
planning environment under a fixed interaction budget (24 turns); we vary the compression
ratio, contrast a dropping operator with a fact-preserving operator at equal budgets,
manipulate the availability of dropped state through oracle restoration, and decompose
tool calls into retrieval versus execution. One model is swept across compression
severities; two further models are evaluated at a fixed ratio, across two task regimes.

Retrieval tool calls increase in every one of our six model--regime comparisons and
account for almost all of the added interaction, while execution calls remain
approximately unchanged; five of six retrieval increases remain significant after Holm
correction. Completion, by contrast, does not track this cost: at the pre-specified $5\times$
comparison point its changes are not significant in any of the six cells (all $p \ge 0.125$),
and even for DeepSeek completion becomes significant only at the most aggressive ratio
($10\times$, $p = .016$) --- the cost signal responds at milder compression than completion does.
GPT-5.5 is the sharpest case ---
completion statistically unchanged (80\% $\to$ 85\%, $p = 1.0$) while retrieval roughly triples
(+42.9 calls). Completion significantly degrades only when reacquisition consumes enough
of the interaction horizon to limit task progress, as observed most clearly for DeepSeek
under high compression. Restoring the dropped state removes roughly half of the retrieval cost,
and a fact-preserving operator at the same compression ratio preserves completion while
avoiding most of the extra retrieval.

Using retention interventions as causal probes, we separate \emph{how much} state is retained
from \emph{which} state is retained and \emph{whether the retained content is valid}. Random
selection matches an offline hindsight oracle, while replacing the retained D-state with
semantically irrelevant content increases retrieval by 57\% ($p < 0.001$) with completion
statistically unchanged --- completion-centric evaluation can fail to expose both the
magnitude and the behavioral consequence of a cost-changing intervention.

We do not claim compression has a fixed cost: in a second environment (ALFWorld) the same
operator produces no retrieval surge, so the reacquisition signature is
environment-dependent rather than an intrinsic consequence of shortening context. In our
controlled setting, the hidden interaction cost of compression emerges when
execution-relevant state becomes absent and must be reacquired; completion alone can miss
that cost, while the magnitude --- and even the presence --- of the cost depends on model,
task, state, and environment. We provide a tool-level diagnostic for a cost that
completion-centric evaluation does not expose.
\end{abstract}

\section{Introduction}
\label{sec:intro}

Long-horizon agents accumulate trajectories that quickly exceed any context window, so
modern runtimes compress --- via summarization, sliding windows, or retrieval filters --- as
a matter of course. A common evaluation view is that compression is successful when it
preserves task performance while reducing context cost; failure-aware compressors can
match full-context performance on many tasks (ACON), and we find even a deterministic
extractive summary to be near-lossless at $5\times$ compression. Existing evaluations ask
whether task performance survives. We ask a complementary question: what runtime cost can
accumulate before that performance changes? Task completion is an incomplete measure of
the runtime cost of context compression --- it does not by itself reveal the interaction
cost incurred when the agent must reacquire compressed-away state.

``Near-lossless'' measures \emph{retention}: whether the surviving context still contains the
needed facts. It does not measure \emph{reacquisition}: what the agent does when a piece of
execution-relevant state is genuinely absent --- re-querying the environment for task
graphs, hidden constraints, resource occupancy --- and what that costs in the agent's own
currency. Task completion is bounded by the interaction horizon, while tool calls
provide a direct measure of the interaction cost spent on reacquisition. Whether
compression becomes costly is therefore not determined by retained content alone;
it also depends on what the agent must do to reacquire the state that is no longer
available. Completion is a blunt instrument for this cost: it is co-determined by the
interaction horizon and can remain unchanged while the reacquisition burden grows.

Prior work on context compression asks four questions: \emph{what to keep} (methods: ACON;
Memento; RE-TRAC), \emph{whether information survives or is recoverable} (Context Codec's
round-trip guarantees; Reclaim's black-box recovery), \emph{whether surviving information is
used} (the ``lost in compaction'' attention bottleneck), and \emph{which context-management
strategy performs best on capability and efficiency jointly} (aggregate efficiency
evaluation; Less Context, Better Agents; agent-evaluation surveys call cost an
under-measured axis of LLM-agent assessment). These lines evaluate what compression
preserves or achieves. None measures what the agent does when state is actually gone.
We ask a fifth question: \textbf{what does it cost the agent to re-acquire what was dropped,
and is that cost visible to standard metrics?} This is the gap we enter --- a controlled
demonstration that the two can diverge.

We introduce a controlled measurement protocol for the reacquisition cost of context
compression. The setup fixes a single interaction horizon (24 turns); varies the
compression ratio via a sliding-window operator; spans two task regimes that differ in
how much execution-relevant state must be discovered at runtime; \emph{manipulates the
availability of execution-relevant state} through controlled oracle restoration ---
injecting specific dropped state back into the context --- and decomposes the agent's tool
calls into \emph{retrieval} (re-acquiring external state) versus \emph{execution} (performing the
task's operations). We run the protocol across three models --- DeepSeek (deepseek-v4-flash), Qwen
(qwen3.7-plus), and GPT-5.5 --- to separate what holds across models from what is model-specific (DeepSeek at
a full severity sweep, Qwen and GPT-5.5 at a fixed $5\times$ ratio).

Our contributions are:

\begin{enumerate}
\item \textbf{Evaluation non-identifiability under context compression.}
  Task completion is not an identifying projection of interaction cost: we formalize this
  as a non-identifiability result and realize it under controlled intervention. GPT-5.5
  pays the largest reacquisition cost in our study while its completion is not detected
  as changed ($80\% \to 85\%$, $p = 1.0$; retrieval $21.0 \to 63.9$, $p = .002$); and in a retention
  intervention (Sec.~\ref{sec:retention}), injecting semantically irrelevant state raises retrieval by 57\%
  ($p < 0.001$) while completion is statistically unchanged ($p = 0.41$). Completion is
  blind not only to the \emph{magnitude} of interaction cost, but also to its \emph{behavioral
  consequence} --- completion did not expose the large cost difference induced by the
  intervention.

\item \textbf{A mechanistic decomposition of the hidden cost.} We decompose the reacquisition
  burden along a recoverability axis: externally queryable task state (D) versus
  history-dependent state (R), with a re-query loop in which the loss of R inflates
  defensive re-querying of D. Retrieval calls increase in all six of our model--regime
  comparisons and account for almost all of the added interaction; restoring dropped
  state removes roughly half the retrieval cost; and across a retention-budget axis the
  reduction tracks the retained coverage of D-state (a dose--response within a fixed
  digest format).

\item \textbf{Retention interventions as causal probes: the behavioral relevance of retained
  state.} We use a family of retention interventions to separate \emph{how much} state is
  retained from \emph{which} state is retained and \emph{whether the retained content is valid}.
  Fine-grained selection among real, task-relevant atoms has limited marginal value in
  our setting --- random selection matches an offline hindsight oracle --- while replacing D
  content with semantically irrelevant state sharply increases retrieval. The content
  effect replicates across models at the loose budget (GPT-5.5, Sec.~\ref{sec:crossmodel}) while its
  magnitude and budget-gating remain model-specific. State content is behaviorally
  load-bearing, but this is visible only when the digest occupies a substantial share of
  the context window (a content $\times$ budget interaction). An external probe in ALFWorld
  (Sec.~\ref{sec:alfworld}) bounds the whole signal: where the relevant state can be re-observed directly,
  the same operator produces no retrieval surge --- so the hidden interaction cost is
  conditional on what execution-relevant state becomes unavailable and must be
  reacquired, and completion metrics do not expose it.
\end{enumerate}

Across three models and two task regimes we further characterize the \emph{boundary} of the
cost signal: reacquisition cost consistently increases, while its translation into
completion loss varies substantially --- DeepSeek degrades at high compression, Qwen is
largely insensitive, and GPT-5.5 does not. The cost signal is also the earlier one:
retrieval responds at milder compression ($5\times$) before completion responds, which for
DeepSeek occurs only at $10\times$. A second environment (ALFWorld, Sec.~\ref{sec:alfworld}) bounds the signal from
the other side: there, the same sliding compression produces no retrieval surge at all,
so even the \emph{presence} of the cost is conditional --- it emerges when execution-relevant
state must be reacquired, and completion metrics do not expose it.

This paper is not another compression benchmark, nor a claim that compression uniformly
hurts agents --- our own data rule that out. It is a \emph{controlled empirical demonstration of
an evaluation blind spot}: a measurement protocol showing that, in this setting, the
reacquisition cost is real, decomposable, and model-dependent --- and that
completion-centric evaluation can miss it, even under an intervention that changes the
cost sharply. We do not propose a general theory of agent evaluation; we exhibit one
mechanism by which completion can under-report compression's cost, and we give runtime
designers a diagnostic for it. For runtime designers, the actionable quantity is not
``which compression ratio is safe'' but ``which execution-relevant state, if dropped, would
the agent spend its budget re-acquiring?'' --- and our results identify externally queryable
task state as a major source of that cost, with the retention-budget results
(Sec.~\ref{sec:retention}) qualifying how much the digest's content --- as opposed to its mere presence --- is
responsible for reducing it.

\section{Related Work}
\label{sec:related}

\textbf{Category map.} Prior work on context compression spans four categories. (i)
\emph{Compression methods} ask how to compress better --- what to keep and when to compact
(ACON; Memento; VISTA; RE-TRAC; LLMLingua) --- and optimize for agent
performance. (ii) \emph{Recoverability evaluation} asks whether information survives or can
be recovered (Context Codec's round-trip guarantees; Reclaim's black-box recovery) --- measuring retention as a property of the compressed artifact. (iii)
\emph{Long-context utilization} asks whether surviving information is actually used, showing
that compaction can fail through attention rather than deletion (``lost in compaction'').
(iv) \emph{Aggregate efficiency evaluation} reports completion together with aggregate cost ---
tokens, latency, runtime --- when comparing context-management strategies (Less Context,
Better Agents), and agent-evaluation surveys and benchmarks increasingly argue that
success alone is insufficient and must be paired with cost (Yehudai et al.; CostBench).
Each of these lines evaluates what compression preserves, recovers, or achieves. None
provides a diagnostic for the interaction cost that completion-centric evaluation does
not expose --- where the added interaction goes, and whether completion tracks it. This
paper is that diagnostic: a controlled protocol, in a fixed setting, that measures the
reacquisition cost and tests whether completion exposes it.

\textbf{Evaluation metrics for agents, and the insufficiency of completion.} A separate line of
work asks what the \emph{right metric} for an agent is. Production experience shows completion
alone is a blunt instrument: agents in the wild re-query tools and re-fetch
already-computed state, and completion reads the final outcome, not the trajectory behind
it. Formalizations make the insufficiency precise from different directions: completion
rate compresses process-level variance --- the ``completion fallacy'' (Hou et al.) --- binary
pass/fail obscures meaningful differences among failing agents, motivating graded
reliability and progress-based metrics (Khanal et al.; AgentBoard), and bounded-horizon
outcome decomposition reveals horizon-dependent trade-offs (the Verifier Tax). A parallel
line formalizes the cost side: the tool-calling protocol itself carries an additive
accuracy tax (Zhang et al.), cost-optimal planning is benchmarked through cost gaps
(CostBench), and cost-per-return frameworks make effectiveness--efficiency trade-offs
explicit (Bogdanov et al.). Two observations bracket, but do not fill, the gap we target.
These insufficiency arguments concern what completion fails to reveal about process quality
or the outcome --- not about the runtime cost of context compression; and the cost-accounting
frameworks price architecture or protocol choices under an intact context, without
connecting cost to state that compression dropped. On the compression side, reacquisition
cost is recognized operationally --- tokens-per-task accounts for re-fetching (Factory),
truncation rots recall (context-clock), and token reduction does not reliably reduce billed
cost (Weinberger \& Hozez) --- but only observationally, without causal attribution. None
defines a black-box, model-agnostic measure of how much interaction budget an agent spends
to reacquire state that generic compression removed, nor manipulates that state's
availability causally. Our protocol supplies exactly this measurement (Proposition~1,
Sec.~\ref{sec:nonident}).

\textbf{Closest evaluation-perspective contrasts.} Table~\ref{tab:contrasts} summarizes the four closest
evaluation-perspective works. None measures the interaction budget an agent spends to
reacquire state that a compression operator dropped, nor manipulates state availability
causally; our protocol does.

\begin{table}[t]
\centering
\small
\caption{Closest evaluation-perspective contrasts.}
\label{tab:contrasts}
\begin{tabular}{@{}>{\raggedright\arraybackslash}p{2.0cm}>{\raggedright\arraybackslash}p{1.9cm}>{\raggedright\arraybackslash}p{1.9cm}>{\raggedright\arraybackslash}p{1.9cm}>{\raggedright\arraybackslash}p{2.1cm}>{\raggedright\arraybackslash}p{2.1cm}@{}}
\toprule
\textbf{Work} & \textbf{Measures} & \textbf{Manipulates} & \textbf{Decomposes} & \textbf{What completion can miss} & \textbf{Our difference} \\
\midrule
\textbf{CostBench} & cost-optimal planning; cost gaps & tool costs \& availability (blocking events) & cost by event type & cost-suboptimal but valid plans & intact context; we price reacquiring dropped state \\
\textbf{Compress\-Agent} & reliability \& success vs compression & compression retention level & failures: tool-execution vs reasoning & tool failures precede success collapse & no causal state manipulation; we add oracle attribution \\
\textbf{Token Red.\,$\neq$\,Cost} & billed cost; token vs cost deltas & token-reduction strategies & cost by billing component (cache $\approx$87\%) & token savings $\neq$ cost savings & provider-specific billing; we use a model-agnostic budget \\
\textbf{Tool-Use Tax} & accuracy tax of the tool protocol & tool availability (no-op / oracle probes) & accuracy gap: $\Delta$cmp + $\Delta$frc + $\Delta$sty & protocol tax swamps tool gains & accuracy over intact context; we meter cost of lost state \\
\bottomrule
\end{tabular}
\end{table}

\textbf{Closest contrasts.}

\begin{itemize}
\item \textbf{Lost in Compaction} shows compaction fails through an attention bottleneck rather
  than deletion (``grep-LLM gap'': keyword search finds 82--93\% of facts while the LLM
  recalls 0--7\% in the compacted zone). Our mechanism targets a distinct failure mode and
  is experimentally separable in our controlled setup: our sliding condition \emph{deletes}
  old turns, so the dropped state is genuinely absent and the agent's observable response
  is to re-fetch it (retrieval calls increase 2--3$\times$; execution calls do not); the oracle
  condition confirms this attribution by restoring state and removing much of the
  retrieval cost. We do not contest attention-dilution accounts --- they predict our
  observation that GPT-5.5 tolerates compression --- but we isolate a second, independent
  cost that attention-centric accounts do not capture, because for them the text is still
  present. Both mechanisms can coexist; we measure the reacquisition side.

\item \textbf{Context Codec} formalizes codec-level round-trip recoverability of compressed
  representations. Their recoverability is a representation property (can the encoder +
  decoder round-trip?); ours is a black-box agent-behavioral property (can the agent
  recover under a bounded budget?). Same word, different layer --- complementary.

\item \textbf{Reclaim Evaluation} tests recovery of correctable state from a degraded single-carried
  memory (``a lossy memory is worse than an empty one''). We measure the \emph{cascading reacquisition cost
  inside the agent loop} and manipulate recoverability directly via oracle injection.

\item \textbf{Plans Don't Persist} probes latent hidden states to show internal plans don't
  survive compression. White-box probing of internal state vs our black-box observation
  of environment state (ERS); we make no claim about latent reasoning.

\item \textbf{Memento} compresses the KV cache within a single inference. We study the cross-turn
  agent context --- the trajectory persisting across tool calls --- and the interaction
  budget spent re-acquiring environment state.

\item \textbf{ACON} shows failure-aware compression can match full-context performance. Their
  target is ``compression should not hurt''; we reproduce a near-lossless summary in our
  setup, and show the divergence appears only when execution-relevant state is dropped
  (sliding), consistent with our claim that compression quality depends on which
  execution-relevant state is preserved.

\item \textbf{VISTA} provides lossless archival recovery of exact bytes (bulky blocks archived as
  external payloads with stable handles and restored on demand). We show that exercising such
  recovery is itself a cost continuum under a hard budget: even recoverable task-graph state costs
  real turns to re-fetch, and that cost can become consequential for performance under a bounded
  interaction horizon.

\item \textbf{Less Context, Better Agents} shows that pruning plus summarization can improve
  completion while cutting tokens and runtime in an enterprise tool-use workflow. Their
  evaluation is a configuration-choice question --- which context-management strategy
  performs best --- reported on completion, tokens, and runtime jointly, alongside a
  failure taxonomy. We ask a different question at a finer granularity: where the
  additional interaction goes (retrieval versus execution), and whether completion
  changes even when that interaction cost does. Our contribution is a decomposition and
  a diagnostic, not a competing configuration choice; the two are complementary.

\item \textbf{CompressAgent} independently observes the same ordering we measure: tool-execution
  failures appear at milder compression, while task success collapses only under
  aggressive compression. Where they study compression reliability at scale, we isolate
  the mechanism --- state reacquisition --- under a controlled budget and intervene on it
  with oracle restoration.

\item \textbf{Rate-distortion / memory surveys} explicitly flag a gap: agent-level repeated
  compaction is barely measured and no benchmark shares a budget axis across layers. We
  provide exactly the missing controlled experiment --- operator $\times$ regime $\times$ model under a
  fixed interaction budget.
\end{itemize}

Prior work on context compression asks \emph{what to keep}, \emph{how well compression preserves
information}, \emph{whether surviving information is used}, and \emph{which strategy wins on the
aggregate outcome}. We ask a fifth question --- \textbf{what it costs the agent to re-acquire
what was dropped, and whether completion metrics expose that cost} --- and show that this
cost \textbf{can rise substantially while completion stays unchanged}:

\begin{quote}
\emph{``Compression is not costly merely because information is lost; it becomes costly when
the state that was dropped must be re-acquired through additional interaction under a
bounded budget.''}
\end{quote}

\section{Method}
\label{sec:method}

\begin{figure}[t]
\centering
\includegraphics[width=0.95\textwidth]{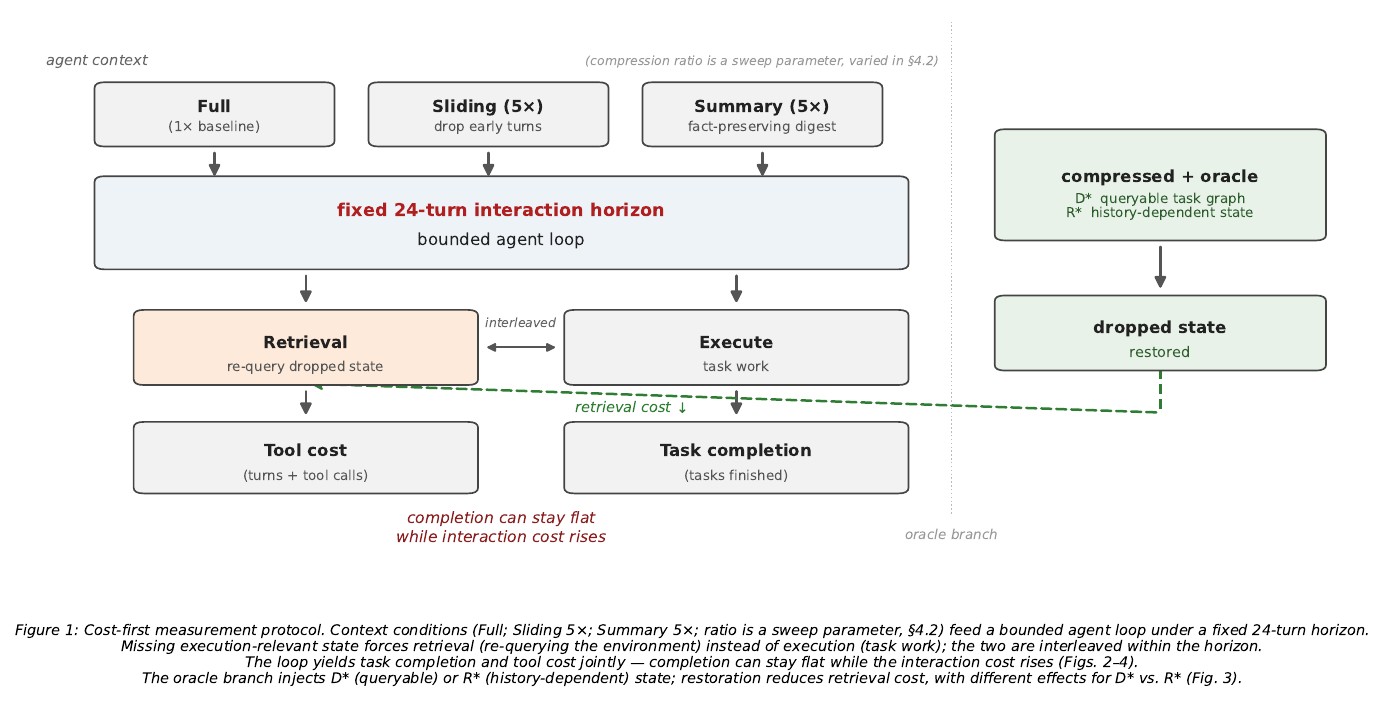}
\caption{Measurement protocol overview: full / fact-preserving vs.\ sliding (compressed)
context; missing execution state diverges into retrieval (reacquisition) vs.\ execution
(task work); a parallel oracle branch (compressed + $D^*$/$R^*$ $\to$ injected dropped state $\to$ retrieval cost drops)
makes recoverability a manipulated variable.}
\label{fig:overview}
\end{figure}

\textbf{Figure~\ref{fig:overview}} (measurement protocol overview): an agent context feeds either a full /
fact-preserving context or a compressed (sliding) context; missing execution state
diverges into \emph{retrieval} (reacquisition) versus \emph{execute} (task work), which together
consume the 24-turn horizon and determine completion / tool cost. A parallel oracle
branch (compressed + $D^*$/$R^*$ $\to$ dropped state injected $\to$ retrieval cost drops) shows
recoverability is a manipulated variable, not a passive observation.

\subsection{Evaluation outcomes, projections, and non-identifiability}
\label{sec:nonident}

We formalize the evaluation object this paper measures. For a context-management condition
$m$ we record a triple of measured outcomes.

\begin{quote}
\textbf{Definition 1 (Evaluation outcome).} $Y(m) = (Q(m),\, C_R(m),\, C_E(m))$, where $Q$ is the task
completion rate, $C_R$ the number of retrieval tool calls, and $C_E$ the number of execution
tool calls, all recorded under a fixed interaction horizon. We interpret $C_R$ as reflecting
state re-acquisition effort and $C_E$ as reflecting task work. We refer to $C = C_R + C_E$ as
the \emph{tool-level interaction cost} --- the agent's budgeted currency, not a claim about
wall-clock or monetary cost.
\end{quote}

Completion-centric evaluation applies a projection $\pi_Q(Y) = Q$ to this triple. We ask when
such a projection can preserve the ordering of strategies by interaction efficiency. Let
$\sim_Q$ denote that the pre-specified paired test (paired Wilcoxon signed-rank with Holm
correction; Sec.~\ref{sec:setup}) does not detect a completion difference between two conditions. A
completion-only evaluation can preserve cost-aware ordering only if $m_1 \sim_Q m_2$ implies the
two conditions also do not differ materially in interaction cost.

\begin{quote}
\textbf{Proposition 1 (Non-identifiability of completion-only evaluation).} The completion-only
projection $\pi_Q$ discards the interaction-cost dimensions of $Y$. Distinct outcome states with
different interaction costs can therefore map to the same completion value. Under our
protocol, this non-identifiability is realized empirically by conditions whose completion
differences are not detected by the pre-specified paired test while their retrieval costs
differ substantially.
\end{quote}

\emph{Empirical counterexample.} Full vs.\ Sliding ($5\times$) on GPT-5.5 under High-IR: completion
$80\% \to 85\%$ (paired $p = 1.0$; 95\% CI $[-6, +21]$ pp, containing zero), while retrieval
$21.0 \to 63.9$ (paired $p = .002$; Cohen's $d = 2.07$; 95\% CI $[+31, +55]$, excluding zero). A
completion-only evaluator would weakly prefer Sliding (85\% vs.\ 80\%), despite a roughly
threefold increase in retrieval cost.

Note the claim is not that the two conditions are equivalent; it is that completion-only
evaluation is blind to the cost dimension. At the metric level, any evaluation that
projects the outcome to completion alone discards the interaction-cost dimensions by
construction; the counterexample above is an empirical instance in which this discarded
dimension is large and operationally relevant (cost difference carries $d = 2.07$, while the
completion difference is not detected). The direction of the completion change is
immaterial: even if completion improved under compression, a completion-only evaluator
would remain blind to the concurrent cost increase.

\textbf{Cost-aware dominance.} We say $m_1$ dominates $m_2$ if $Q(m_1) \ge Q(m_2)$ and $C(m_1) \le C(m_2)$, with
at least one inequality strict --- a two-dimensional Pareto relation on $(Q, C)$.
Completion-only evaluation cannot detect it, since $\pi_Q$ collapses the cost dimension. The
cleanest instance is Summary vs.\ Sliding at $5\times$ on DeepSeek High: Summary achieves higher
completion (83\% vs.\ 72\%) with lower interaction cost (39.0 vs.\ 71.4). A completion-only
evaluator correctly prefers Summary but cannot quantify how much better: the 11pp
completion gap understates the 32.4-call interaction-cost gap. The decomposition $C = C_R +
C_E$ attributes this cost gap almost entirely to retrieval (19.5 vs.\ 55.1; execution 19.5
vs.\ 16.3).

\textbf{Completion-insensitivity region (descriptive).} We use this term descriptively for
(compression ratio, model) pairs where the retrieval cost increase under compression,
$\Delta C_R = C_R(\text{compressed}) - C_R(\text{Full})$, is statistically detectable while the paired test does
not detect a completion change (Cohen's $d \ge 0.5$ as a secondary descriptive criterion).
This is a descriptive characterization with conventional thresholds --- not an equivalence
test, and not a proposed metric; the term denotes an empirical pattern rather than proof of
completion insensitivity. In our data (Fig.~\ref{fig:phasediag}a), GPT-5.5 High at $5\times$ lies inside;
DeepSeek leaves the region at $10\times$, where completion degradation becomes significant; Qwen
High sits at the boundary, with retrieval only weakly elevated and completion modestly
declining.

\subsection{Agent and task environment}
\label{sec:env}

We study a minimal tool-using agent operating on a deterministic project-planning
environment, which we refer to as \textbf{IRBench}. The environment exposes \textbf{10 tasks} and a small set of \textbf{resources} (each
with capacity one). Tasks carry hidden execution constraints that are only revealed
through interaction: ordering rules, resource holds with release times, and
fail-until-success counts. The agent has four tools: \texttt{get\_task\_info}, \texttt{check\_dependency},
and \texttt{query\_resource} (all \emph{retrieval}: they obtain environment state), and \texttt{execute}
(\emph{execution}: it performs a task and reveals constraints). The retrieval/execution split
is central: it lets us measure reacquisition cost directly, without recourse to CoT or
latent reasoning.

The environment is \textbf{deterministic}: a seed fully determines the world instance, so any
condition can be replayed on the identical task set. We generate two task regimes by
varying the density of hidden, execution-relevant facts while keeping static structure
comparable (verified by environment self-checks): \textbf{High-IR} (constraints revealed only
during execution, costly to lose once dropped) and \textbf{Low-IR} (multi-hop dependency
chains, but state re-derivable from the public task graph). Concretely, a task may fail
because it must follow another task, or because a resource remains held until a later
event. The task graph and current resource state can be queried again at any time; a
previously revealed failure condition or fail-until count, by contrast, exists only in
the execution history.

The agent receives a fixed system prompt (identical across all conditions) instructing it
to discover prerequisites, respect constraints, and complete all ten tasks within a
\textbf{fixed interaction horizon of 24 turns}; each turn is one model call plus any tool
results. Completion is judged by how many of the ten required tasks are finished when the
horizon ends. We additionally compute an auxiliary \textbf{tool-budget statistic} ($\alpha \times$ the
reference tool count, $\alpha = 2.0$); this is a \emph{diagnostic}, not a termination criterion --- the
agent is never cut off by it.

\subsection{Context conditions: operators and oracle interventions}
\label{sec:conditions}

We isolate the effect of compression by holding the task instance, system prompt, tool
schema, and model fixed, and varying \textbf{only} the context delivered to the model. All
compression conditions are applied online, at the same point in the agent loop, before
each model response is generated --- the context is rebuilt on every turn, so the operator
sees exactly the trajectory accumulated so far. Turns are grouped into atomic units (an
assistant message plus its tool results), so the context is always well-formed for the
API.

\begin{table}[t]
\centering
\small
\caption{Context conditions.}
\label{tab:conditions}
\begin{tabular}{@{}>{\raggedright\arraybackslash}p{2.2cm}>{\raggedright\arraybackslash}p{13.2cm}@{}}
\toprule
\textbf{Condition} & \textbf{Context} \\
\midrule
\textbf{Full} & the complete trajectory (baseline) \\
\textbf{Sliding} & a token budget of $r^{-1} \times$ the trajectory ($r = 1.7\times$, $2.5\times$, $5\times$, $10\times$); greedily keep the most recent turns within budget; \textbf{drop} early turns outright \\
\textbf{Summary} & the same budget, but dropped early turns are replaced by a deterministic fact digest \texttt{[Summary of earlier trajectory --- observed facts]} (extractive; keeps observed state facts, discards the model's reasoning text); zero extra LLM calls \\
\textbf{Oracle} & Full or Sliding context with a specific state string \textbf{injected as a trailing message} (same position in every condition) \\
\bottomrule
\end{tabular}
\end{table}

Sliding is \emph{deletion} (not summarization), so dropped state is genuinely absent --- this
tests ``absent $\to$ reacquire,'' not ``present but ignored.'' Summary shares the same token
budget, making the operator contrast fair; extractive mode keeps out summarizer-quality
confounds --- our extractive summary preserves the observed state facts represented in the
digest while discarding the trajectory's reasoning text; it is a designed control, not a
general summarizer. Oracle injection is a trailing message in a fixed position, leaving
the system prompt untouched, preserving the same-prompt commitment across conditions.

The oracle supplies exactly the execution-relevant state that sliding drops, split into
two recoverability classes: \textbf{$D^*$} --- externally queryable state (the task graph:
prerequisites, resource occupancy), recoverable in principle with more tool calls; and
\textbf{$R^*$} --- history-dependent state (revealed constraints, failure counts, held resources),
not recoverable from any single public query. These are recoverability projections, not a
data-versus-reasoning dichotomy. This gives us a manipulated independent variable --- the
availability of dropped state --- allowing us to ask: \emph{if the lost state were restored, how
much of the compensatory retrieval disappears?}

\subsection{Protocol and analysis}
\label{sec:setup}

For each condition and model we run the same 10 seeds (10 task instances per seed, 100
task-level observations per condition); the retention-intervention family of
Sec.~\ref{sec:retention} uses an extended \textbf{20 paired seeds} to bind the selection null
(Sec.~\ref{sec:selection}). Statistics are \textbf{per-seed paired} (same seed
under both conditions): Wilcoxon signed-rank tests and bootstrap 95\% confidence intervals
on per-seed differences. We pre-specify the primary
comparisons (Full vs.\ Sliding on task completion, tool calls, and retrieval calls). For
multiple-comparison control, the primary family is the six retrieval comparisons (three
models $\times$ two regimes); we apply Holm--Bonferroni correction at $\alpha = 0.05$, after which five
of six remain significant. Remaining pairwise tests are exploratory. Models: \textbf{DeepSeek}
(deepseek-v4-flash), \textbf{Qwen} (qwen3.7-plus), and \textbf{GPT-5.5}. All were evaluated without an
explicit extended-thinking mode, at temperature 0.3, to keep the observable interaction
protocol as comparable as possible across providers. Each model is served through an
OpenAI-compatible interface.

\subsection{Retention interventions as causal probes}
\label{sec:retention}

The oracle result (Sec.~\ref{sec:oracle}) shows that restoring dropped state reduces reacquisition cost.
This raises a question about \emph{granularity}: at what level does retained state matter to
the agent's interaction cost --- how much is retained, which atoms are retained, and whether
the retained content is valid and task-relevant? We address it with a family of retention
interventions, applied online at a fixed compaction point (turn 10, $t_c$), each taking
the observed trajectory up to $t_c$, extracting the set of observed state atoms, and
re-rendering a compressed prefix under a \textbf{digest budget} $B \in \{265, 100, 50\}$
tokens --- an independent resource axis that does not co-vary with the compression ratio.
Every digest is injected at the same position with the same \texttt{[State retention digest]}
format, followed by the most recent turns within the total budget. Atoms are split into
the D/R classes of Sec.~\ref{sec:conditions}. We log the retained coverage per class ($\mathrm{cov}_D$,
$\mathrm{cov}_R$) and the actual rendered digest size, so the requested budget and the
delivered digest are both reported.

\begin{table}[t]
\centering
\small
\caption{Retention interventions.}
\label{tab:interventions}
\begin{tabular}{@{}>{\raggedright\arraybackslash}p{2.6cm}>{\raggedright\arraybackslash}p{12.8cm}@{}}
\toprule
\textbf{Intervention} & \textbf{Selection rule} \\
\midrule
\textbf{Sliding} & recency baseline; no digest (recent turns only) \\
\textbf{RAR-D} & retain D-type atoms only \\
\textbf{RAR-R} & retain R-type atoms only \\
\textbf{RAR-All} & retain all atoms in observation order ($\approx$ Summary control) \\
\textbf{RAR-TypeAware} & retain all atoms, R-first (recoverability-prioritized) \\
\textbf{Random} & random subset of atoms (seeded) \\
\textbf{Recent} & most-recently-observed atoms \\
\textbf{Hindsight} & top-$k$ by \emph{offline} future re-access counts (reference) \\
\textbf{D-Irrelevant} & as TypeAware (R first), but D-budget filled with fabricated out-of-universe atoms \\
\bottomrule
\end{tabular}
\end{table}

The candidate universe is strictly online: the atoms observed before $t_c$, so no
condition sees future observations. \textbf{Hindsight} is an offline reference, not an upper
bound; we call it an \emph{offline hindsight oracle under the reference trajectory}. The
recoverability-prioritized ranking (R before D) is a pre-registered hypothesis being
\emph{tested}, not an assumed ordering --- the Random and Recent controls exist precisely to
test whether ranking beats selection that ignores it. The digest budget $B$ is the
independent variable of a budget sweep (all interventions at all three budgets).

\textbf{Pre-registration and the D-Irrelevant control.} We pre-registered the primary
comparison (RAR-D vs.\ Sliding on retrieval cost, criterion $\Delta C_R \le -30\%$ with a
95\% CI excluding zero) and the condition matrix above. As reported in Sec.~\ref{sec:retention}, the
fine-grained selection null (Random $\approx$ Hindsight) emerged at $N=20$, prompting a
mechanistic follow-up: the \textbf{D-Irrelevant} intervention, which holds the digest format,
position, budget, and retained R content fixed and replaces only the D atoms with
fabricated out-of-universe state (identifiers outside the task/resource universe, so no
entity can be confused with the current world). Its judgment criterion
($\Delta C_R = C_R[\text{D-irrelevant}] - C_R[\text{TypeAware}]$, paired 95\% CI) was
fixed before the intervention ran. We report it transparently as a post-hoc sequential
diagnostic rather than a pre-registered primary.

\section{Experiments}
\label{sec:experiments}

\subsection{Setup}
\label{sec:exp-setup}

We follow the protocol of Sec.~\ref{sec:method}. Unless stated otherwise, ``completion'' is the fraction of
the ten required tasks completed within the fixed 24-turn horizon; ``tools'' is the per-run
tool-call count, decomposed into retrieval and execution calls. Each condition runs on
10 seeds, analyzed per-seed paired (Wilcoxon signed-rank; bootstrap 95\% CI); primary
comparisons are Full vs.\ Sliding on completion, tools, and retrieval calls.

\begin{figure}[t]
\centering
\includegraphics[width=0.95\textwidth]{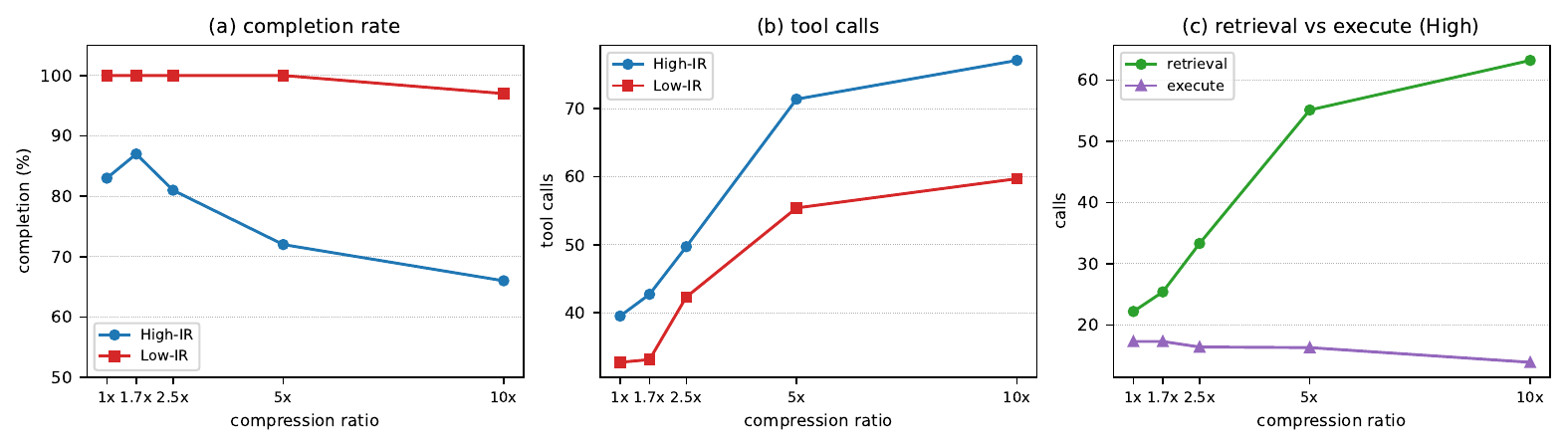}
\caption{Compression raises reacquisition cost before completion may degrade. Panel A:
completion vs.\ compression ratio; Panel B/C: tool / retrieval--execution decomposition.}
\label{fig:sweep}
\end{figure}

\subsection{Compression raises reacquisition cost before completion may degrade (Figure~\ref{fig:sweep})}
\label{sec:sweep}

\textbf{Sweep, DeepSeek (High).} Varying compression severity with the sliding-window operator
($r \in \{1, 1.7, 2.5, 5, 10\}$) on High-IR:

\begin{table}[t]
\centering
\small
\caption{DeepSeek, High regime: severity sweep.}
\label{tab:sweep}
\begin{tabular}{@{}rrrrrr@{}}
\toprule
ratio & completion & tools & retrieval & execute & turns \\
\midrule
$1\times$ (Full) & 83\% & 39.5 & 22.2 & 17.3 & 22.9 \\
$1.7\times$ & 87\% & 42.7 & 25.4 & 17.3 & 23.3 \\
$2.5\times$ & 81\% & 49.7 & 33.3 & 16.4 & 22.8 \\
$5\times$ & 72\% & 71.4 & 55.1 & 16.3 & 23.1 \\
$10\times$ & 66\% & 77.1 & 63.2 & 13.9 & 23.6 \\
\bottomrule
\end{tabular}
\end{table}

Completion is flat through moderate ratios --- the small uptick at $1.7\times$ (87\% vs.\ 83\% at
Full) is within run-to-run noise and consistent with mild compression removing stale or
distracting state, the mechanism behind ``less context, better agents''; it does not
persist --- and declines only at the two most aggressive ratios ($5\times$: $-11$pp, $p = .25$ NS;
$10\times$: $-17$pp, \textbf{$p = .016$}). This is the only completion change in our study to reach
significance, and it does so only at the most aggressive ratio. Retrieval, by contrast,
is already significantly elevated at $5\times$ ($p = .002$): the cost signal responds at milder
compression than completion does (Sec.~\ref{sec:models}). Moderate compression does not yield a monotonic
completion decline; the more stable signal is the accompanying growth in interaction
cost. Tool calls increase
monotonically from $2.5\times$ onward, with paired differences significant under our
pre-specified test: tools rise from 39.5 to 77.1 (\textbf{+37.6 at $10\times$, $p < .01$}). The
decomposition shows \emph{where} the extra interaction goes: retrieval calls rise from 22.2 to
63.2 (\textbf{+41.0}, $p < .01$), while \textbf{execute calls do not increase with severity} (they
fall from 17.3 to 13.9). Compression does not make the agent attempt more task work; it
makes the agent spend more interaction re-acquiring state that was dropped. Across
independent collections
the full-context baseline varied by only a few percentage points (80--86\%), within the
run-to-run variability of this stochastic setting; we use the canonical sweep for all
primary comparisons.

\textbf{Regime contrast (Low).} Under Low-IR the same operator is near-costless in completion
(100\% at every ratio until $10\times$, 97\%) while the tool response is preserved: tools rise
32.7 $\to$ 59.7 (+83\% at $10\times$), retrieval-dominated. Low-IR's state is re-derivable from the
public graph, so the additional reacquisition cost remains small enough to preserve
completion within the evaluated horizon.

\subsection{What is retained matters more than the ratio (operator contrast)}
\label{sec:operator}

The same $5\times$ budget can be spent two ways: sliding (drop old turns) or extractive summary
(compress old turns into observed state facts). Holding budget and model fixed:

\begin{table}[t]
\centering
\small
\caption{Operator contrast at $5\times$, High regime (DeepSeek).}
\label{tab:operator}
\begin{tabular}{@{}lccc@{}}
\toprule
condition ($5\times$, High) & completion & tools & retrieval \\
\midrule
Full & 80\% & 37.4 & 19.0 \\
Sliding & 72\% & 71.4 & 55.1 \\
\textbf{Extractive Summary} & \textbf{83\%} & \textbf{39.0} & \textbf{19.5} \\
\bottomrule
\end{tabular}

\footnotesize
$^\dagger$ Sliding is the $5\times$ condition of the canonical sweep (Sec.~\ref{sec:sweep}); Full and Summary come from a
separate operator-contrast batch. Full-context completion varied 80--86\% across batches
(Sec.~\ref{sec:sweep}), so the Sliding--Summary gap (72\% vs.\ 83\%) exceeds the batch-level spread of a few points.
\end{table}

With the identical compression ratio, the summary operator is near-lossless (completion
83\% $\approx$ full 80\%; tools flat 37.4 $\to$ 39.0), while sliding degrades completion and triples
retrieval. The extractive operator preserves observed state facts while discarding the
trajectory's reasoning text; sliding drops both. This controlled contrast indicates that
\textbf{what survives compression can matter more than the compression ratio itself}. It also
reproduces ACON-style ``compression need not hurt'' --- but only for the operator that
preserves state facts, which is consistent with our claim.

\subsection{Restoring dropped state reduces reacquisition (oracle intervention)}
\label{sec:oracle}

To make the causal link explicit, we inject the exact dropped state back into the sliding
context as a trailing message ($D^*$ = queryable task graph; $R^*$ = history-only
constraints/fail-counts):

\begin{table}[t]
\centering
\small
\caption{Oracle intervention at $5\times$, High regime (DeepSeek).}
\label{tab:oracle}
\begin{tabular}{@{}lccc@{}}
\toprule
condition ($5\times$, High) & completion & tools & $\Delta$ tools vs sliding \\
\midrule
Full & 86\% & 38.8 & --- \\
Sliding & 66\% & 72.9 & --- \\
Sliding + $R^*$ & 78\% & 69.0 & $-3.9$ \\
Sliding + $D^*$ & 80\% & 35.8 & \textbf{$-37.1$} \\
Full + $R^*$ (sanity) & 85\% & 32.1 & --- \\
\bottomrule
\end{tabular}
\end{table}

Restoring $D^*$ removes most of the compensatory retrieval (72.9 $\to$ 35.8 tools) and recovers
most of the completion gap (66\% $\to$ 80\%). Injecting $R^*$ helps completion (+12pp) but leaves
the retrieval cost high (69.0), because $R^*$ is history-only and its absence is what forces
defensive re-querying of D. The sanity condition (full + $R^*$) is harmless (85\% $\approx$ full
86\%), confirming that oracle injection does not distort the baseline.
The intervention provides causal evidence that removing execution-relevant state increases
reacquisition behavior, while restoring state reduces that behavior.

\subsection{The cost generalizes; the outcome does not (three models)}
\label{sec:models}

We repeat the Full vs.\ Sliding ($5\times$) comparison on Qwen (qwen3.7-plus) and GPT-5.5 (High and Low):

\begin{table}[t]
\centering
\small
\caption{Three models, Full vs.\ Sliding ($5\times$), both regimes.}
\label{tab:models}
\begin{tabular}{@{}lrrrrrr@{}}
\toprule
Model & regime & completion Full$\to 5\times$ & $\Delta$ (p) & tools Full$\to 5\times$ & retrieval $\Delta$ (p) & execute $\Delta$ \\
\midrule
DeepSeek & High & 83 $\to$ 72 & $-11$ (0.25) & 39 $\to$ 71 & +32.9 (.002)$^\dagger$ & $-1.0$ \\
Qwen & High & 75 $\to$ 66 & $-9$ (0.13) & 34 $\to$ 37 & +2.9 (.088) & $\approx 0$ \\
GPT-5.5 & High & 80 $\to$ 85 & +5 (1.0) & 39 $\to$ 82 & +42.9 (.002)$^\dagger$ & $\approx 0$ \\
DeepSeek & Low & 100 $\to$ 100 & 0 & 33 $\to$ 55 & +22.6 (.004)$^\dagger$ & +0.1 \\
Qwen & Low & 100 $\to$ 94 & $-6$ (0.25) & 28 $\to$ 36 & +5.3 (.023)$^\dagger$ & +2.7 \\
GPT-5.5 & Low & 100 $\to$ 100 & 0 & 26 $\to$ 49 & +22.4 (.002)$^\dagger$ & +0.7 \\
\bottomrule
\end{tabular}

\footnotesize
$^\dagger$ $p < .05$ after Holm--Bonferroni correction across the six-comparison retrieval family
($\alpha = 0.05$); Qwen High retrieval is the single non-significant cell.
\end{table}

\textbf{The comparison point is pre-specified.} The six-cell matrix uses a single compression
ratio, $5\times$, fixed in the design of Experiment A (fraction = 0.2) --- before the severity
sweep revealed where completion degrades. We evaluate one representative point per model
rather than a full sweep because the three-model matrix runs under a shared API budget;
$5\times$ is the pre-registered main comparison, and DeepSeek's full severity sweep is reported
separately (Sec.~\ref{sec:sweep}). This is not an attempt to avoid a significant completion result:
DeepSeek's completion does become significant at $10\times$ ($\Delta-17$pp, $p = .016$), and that is
consistent with --- indeed predicted by --- our claim. The cost signal (retrieval) is
significant already at $5\times$ ($p = .002$), while completion responds only at the most
aggressive ratio ($10\times$): completion is the less sensitive measure. If one measured only
completion at $5\times$ --- the shallow-compression operating point many production systems run
at --- one would conclude compression is lossless while retrieval is already elevated.

Across our six model--regime comparisons, retrieval tool calls increased in every
comparison and execution calls remained approximately stable in five of six comparisons;
five of six retrieval increases remain significant after Holm--Bonferroni correction
across the pre-specified family ($p = 0.002, 0.002, 0.002, 0.004, 0.023, 0.088$).
Completion is more heterogeneous, and none of its changes reach significance
(all $p \ge 0.125$): DeepSeek's high-compression degradation is suggestive but not
significant ($-11$pp, $p = .25$), Qwen is largely insensitive --- its retrieval response is
also the weakest ($p = .088$), possibly reflecting reliance on inference rather than
re-querying when state is missing --- and GPT-5.5's completion is
statistically unchanged ($p = 1.0$) while it pays the largest reacquisition cost (+110\%
tools, entirely retrieval). The reacquisition pattern is consistent across the models in
our study; its conversion into reduced completion is not. This heterogeneity is the
boundary condition of our claim, not a contradiction. (The cross-model comparisons use
the same task seeds within each model but are not intended as paired comparisons across
models.)

\begin{figure}[t]
\centering
\includegraphics[width=0.95\textwidth]{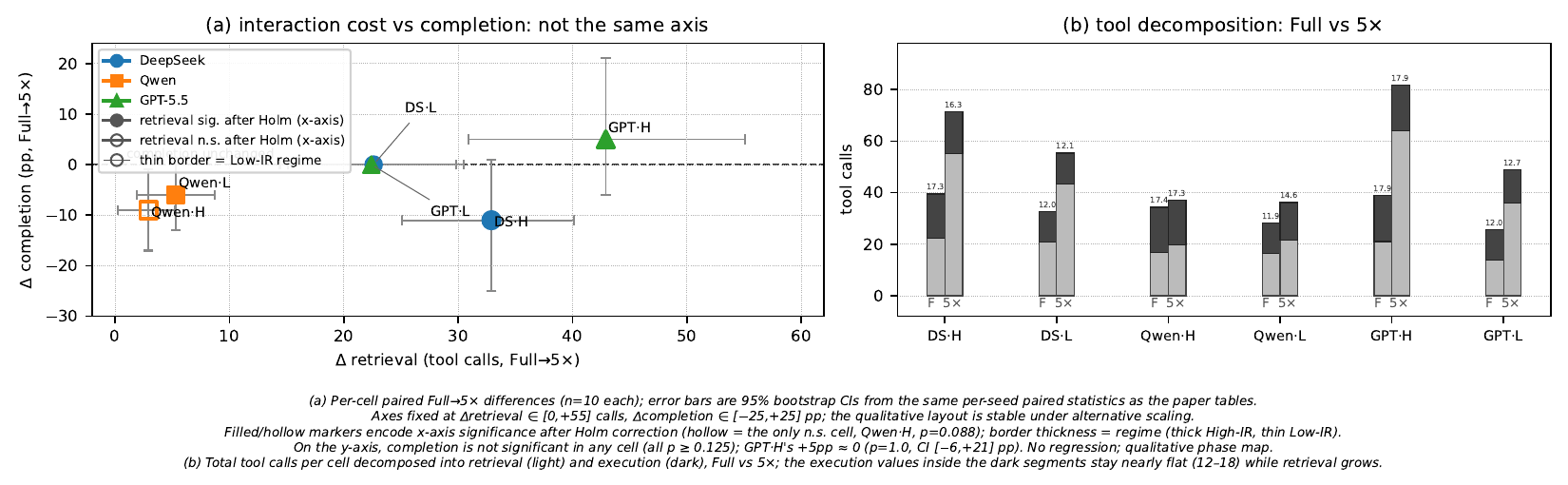}
\caption{Two axes, not one. (a) Phase map of the six model--regime cells in
$(\Delta\text{retrieval}, \Delta\text{completion})$ space; (b) tool decompositions.}
\label{fig:phasediag}
\end{figure}

\textbf{Two axes, not one (Figure~\ref{fig:phasediag}).} Consistent with the dose-response pattern of
Sec.~\ref{sec:sweep}, the six cells occupy a two-dimensional space rather than a single success axis
(Fig.~\ref{fig:phasediag}a): high retrieval growth with completion loss
(DeepSeek High), high retrieval growth with completion stable (GPT-5.5 High, and both
Low cells), and little retrieval growth with a mild completion drop (Qwen). Completion
changes are not significant in any cell (all $p \ge 0.125$), whereas five of six retrieval
increases survive Holm correction --- the two quantities are related but not
interchangeable outcomes. Compression sensitivity is therefore two-dimensional:
interaction cost and task completion occupy distinct axes (no regression is fitted; the
figure is a qualitative phase map). GPT-5.5 High is the sharpest illustration:
the paired test does not detect a completion change ($p = 1.0$; the wide CI
$[-6,+21]$ pp reflects high run-to-run variance at this operating point) while retrieval
roughly triples (+42.9, $p = .002$). Panel (b) shows the same cells as tool
decompositions: the added calls are almost entirely retrieval, with execution
approximately flat.

\begin{figure}[t]
\centering
\includegraphics[width=0.95\textwidth]{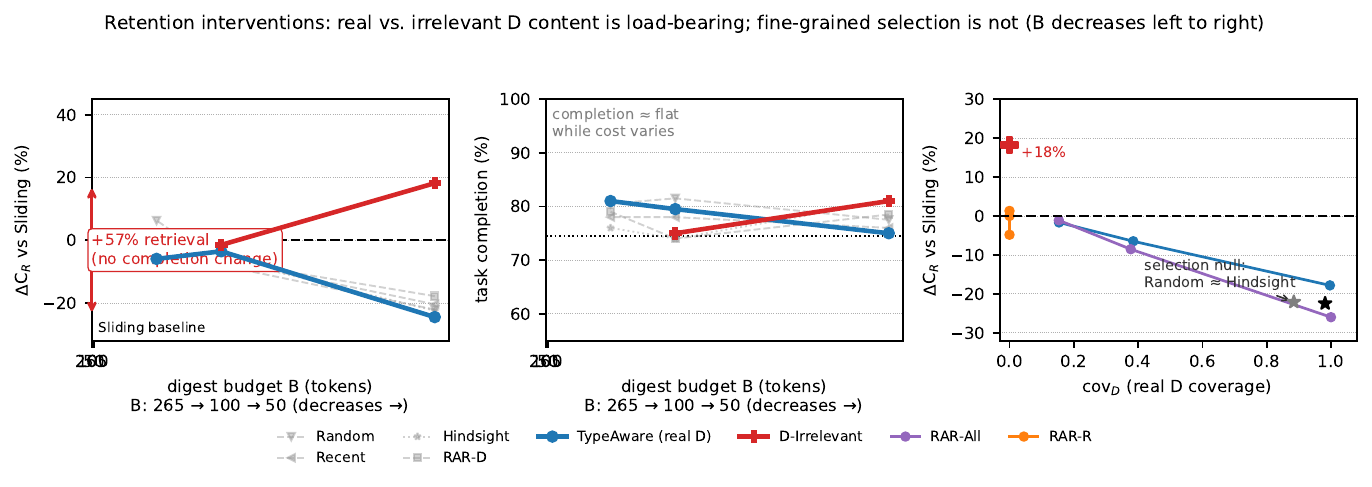}
\caption{Retention interventions. Digest budget $B$ decreases from left to right in
(a)--(b). (a) $\Delta C_R$ vs.\ Sliding (\%): TypeAware (real D) and D-Irrelevant are
highlighted against the faded remaining policies; the $\sim$57\% retrieval gap at $B{=}265$
leaves completion statistically unchanged. (b) task completion ($\approx$ flat across
policies and budgets). (c) coverage dose--response ($\mathrm{cov}_D \to \Delta C_R$) with the
selection null (Random $\approx$ Hindsight) and the D-Irrelevant point at $B{=}265$.}
\label{fig:retention}
\end{figure}

\subsection{Retention interventions reveal two distinct levels of state dependence (Figure~\ref{fig:retention})}
\label{sec:retention-results}

We now report the retention-intervention matrix (Sec.~\ref{sec:retention}) on DeepSeek (High regime,
20 paired seeds; 540 runs total --- the main matrix is 25 runs per seed [sliding at $B=265$
plus eight interventions at three budgets] for 500 runs, plus the D-Irrelevant control at
$B\in\{265,100\}$ for 40 more). The results separate two
questions that the compression sweeps of Sec.~\ref{sec:sweep} conflate: \emph{which} atoms are retained, and
\emph{whether the retained content is valid and task-relevant}.

\subsubsection{Selection null: which atoms are retained does not matter}
\label{sec:selection}

Most structured digests reduced retrieval cost relative to the recency baseline, but the
magnitude did not depend reliably on the selection policy:

\begin{table}[t]
\centering
\small
\caption{Selection null at $B=265$ (DeepSeek, High).}
\label{tab:selection}
\begin{tabular}{@{}lrr@{}}
\toprule
policy ($B=265$) & $\Delta C_R$ vs Sliding & p \\
\midrule
RAR-All & $-26.0\%$ & $<$0.001 \\
RAR-TypeAware & $-24.5\%$ & $<$0.001 \\
Hindsight & $-22.4\%$ & 0.001 \\
Random & $-22.1\%$ & 0.001 \\
Recent & $-20.4\%$ & 0.002 \\
RAR-D & $-17.8\%$ & 0.002 \\
RAR-R & $-4.8\%$ & n.s. \\
\bottomrule
\end{tabular}
\end{table}

Random selection matches the offline hindsight oracle ($-22.1\%$ vs.\ $-22.4\%$; paired
difference +0.10, 95\% CI $[-2.3, +2.3]$, $d \approx 0.02$). Recoverability-prioritized
retention (TypeAware) does not beat Random or Recent on retrieval cost or completion
(all paired comparisons n.s.). Coverage structure holds --- TypeAware keeps
$\mathrm{cov}_R \approx 1.0$ across budgets while the unranked RAR-All digest collapses
$\mathrm{cov}_R$ from 0.30 to 0.00 as the budget tightens --- but this coverage advantage
does not translate into differential cost or completion. R-type state is inherently
compact ($\sim$35 tokens, so the R-only digest is identical at every budget), which is why
even the tightest budget leaves R fully retained.

The paired selection effect is small: with 20 paired seeds, the observed paired
differences and their 95\% CI rule out large selection benefits in this experiment, an order of
magnitude smaller than the content effect below. While a larger $N$ could in principle detect a marginal difference, the practical
conclusion is that fine-grained selection among real, task-relevant atoms has low marginal
value in this setting relative to the coarse content-validity effect.

\subsubsection{Content intervention: the D-Irrelevant control}
\label{sec:content}

The selection null leaves open whether the digest's \emph{content} matters at all, or whether
any D-shaped text occupying the budget is equivalent. The D-Irrelevant control holds the
digest format, position, budget, and retained R content fixed and replaces the D atoms
with fabricated out-of-universe state. At $B=265$ the effect is large and uniform:

\begin{table}[t]
\centering
\small
\caption{D-Irrelevant content intervention at $B=265$ (DeepSeek, High).}
\label{tab:content}
\begin{tabular}{@{}lrrrrr@{}}
\toprule
metric & TypeAware (real D) & D-Irrelevant & $\Delta$ & 95\% CI & p \\
\midrule
$C_R$ (retrieval) & 20.4 & 31.9 & \textbf{+11.5} & [9.2, 13.9] & $<$0.001 \\
tools & 37.6 & 48.2 & +10.7 & [8.1, 13.4] & $<$0.001 \\
execute & 17.2 & 16.4 & $-0.8$ & $[-1.9, +0.2]$ & n.s. \\
Q (completion) & 75.0 & 81.0 & +6.0 pp & $[-2.5, +16.0]$ & 0.41 \\
\bottomrule
\end{tabular}
\end{table}

All 20 seeds show the retrieval increase (range +3 to +23); the increment is entirely in
retrieval, with execute unchanged. Semantically irrelevant state injection is \emph{harmful}:
D-Irrelevant's retrieval exceeds even the no-digest sliding baseline (+18\%, $p = 0.002$).
We report this as a post-hoc sequential diagnostic (see Sec.~\ref{sec:retention}): semantically irrelevant
state injection induces additional verification and retrieval behavior. Whether that
added retrieval reflects semantic irrelevance itself or verification of internally
inconsistent state content is a finer mechanistic distinction we leave for future work;
both readings support the claim that \textbf{state content is behaviorally load-bearing}.

\subsubsection{Budget dependence: content matters when the digest dominates the window}
\label{sec:budget}

The content effect is not budget-invariant. At $B=100$ the same comparison is
null ($\Delta{+0.6}$, 95\% CI $[-2.4, +3.4]$, $p = 0.60$, $d \approx 0.09$): the $\sim$77-token digest is a
minor fraction of the compacted window, and the agent leans on the recent turns, so
real-versus-irrelevant content is behaviorally invisible. At $B=265$ the digest
($\sim$209 tokens) dominates the compressed window and content becomes load-bearing. Content
relevance therefore matters as the interaction \emph{content $\times$ budget}: coarse, valid,
task-relevant D content reduces retrieval at loose budgets, fine-grained selection among
such content does not, and the distinction only becomes behaviorally visible when the
digest occupies a substantial share of the context.

Consistent with this, in our setting recency is a competitive approximation to
recoverability-aware selection under tight budgets, and when the retention budget is
tight, preserving history-dependent (R) state is strongly favored by the intervention
results --- because R is compact enough to retain fully at any budget, and dropping it is
what inflates defensive re-querying of D (the RAR-D cell of Sec.~\ref{sec:selection}).

\subsubsection{Cross-model replication of the content intervention}
\label{sec:crossmodel}

To test whether the content-relevance effect is specific to DeepSeek, we replicate the
key conditions on GPT-5.5 (High regime, $B \in \{265, 100\}$, 10 paired seeds). The
content-relevance effect \textbf{replicates across models at the loose digest budget}, while
its magnitude and budget dependence are model-specific. At $B=265$, the direction matches
DeepSeek exactly (real $<$ irrelevant $<$ sliding on retrieval): GPT-5.5's real-D digest
reduces retrieval relative to sliding ($-9.5$, $p = 0.002$) and its D-Irrelevant control
raises retrieval back up (+7.3 vs.\ real D, $p = 0.006$), while completion is statistically
unchanged (78\% vs.\ 74\%, $p = 0.125$). The content-relevance mechanism is therefore not
model-specific.

Three model-dependent boundaries are worth reporting honestly. First, GPT-5.5's
irrelevant injection does \textbf{not} exceed the raw sliding baseline (27.4 vs.\ 29.6) as it
did in DeepSeek (+18\%); the poisoning intensity is model-dependent. Second, at $B=100$ no
additional content effect was detectable in GPT-5.5 (irrelevant 26.0 vs.\ real 28.8,
$p = 0.125$), consistent with DeepSeek's near-zero effect at the tighter budget. Third, the
mechanism of the GPT response differs in composition: the irrelevant injection reduces
execute calls ($-2.4$, $p = 0.035$) rather than increasing retrieval, suggesting a
model-specific error-correction strategy. We therefore state the result as: the \emph{direction}
of the content effect is reproducible, while its \emph{behavioral expression} is model- and
budget-dependent.

\subsection{External environment boundary: ALFWorld}
\label{sec:alfworld}

To test whether the reacquisition signature is an intrinsic consequence of shortening
context --- or depends on the environment's recoverability structure --- we run the same
Full vs.\ Sliding contrast in ALFWorld, a household instruction-following domain whose
observations are produced by the TextWorld engine. We sample 18 tasks across all six
ALFWorld task types (three tasks per type) and run 5 paired seeds per task (180 runs; all
complete, zero API errors). Retrieval-like actions (look / examine / inventory) are
counted separately from execution-like actions (take / put / clean / heat / cool / use)
and navigation (go).

In our ALFWorld probe, the relevant state could generally be re-observed directly through
the available interaction actions, and we did not observe a retrieval surge under sliding
compression: paired $\Delta$(retrieval-like) $= -0.13$, split 39/35 across seeds (symmetric around
zero); $\Delta$ tools $\approx 0$; and completion did not differ (full 10.0\% vs.\ sliding 8.9\%, with 89/90
paired runs identical). Completion was dominated by the difficulty profile of the sampled
tasks --- only the simplest task type (look\_at) was reliably solvable under this adapter ---
so we treat completion as a secondary outcome in this probe.

\begin{table}[t]
\centering
\small
\caption{External environment boundary.}
\label{tab:envboundary}
\begin{tabular}{@{}llcc@{}}
\toprule
Environment & Compression & $\Delta$ retrieval-like actions & $\Delta$ completion \\
\midrule
IRBench (this work) & Sliding & large positive (2--3$\times$) & model / regime dependent \\
ALFWorld & Sliding & $\approx$ 0 (paired, symmetric) & small / ceiling-bound \\
\bottomrule
\end{tabular}
\end{table}

The reacquisition signature is therefore environment-dependent rather than an intrinsic
consequence of shortening context. In IRBench, dropped task-graph state must be re-queried;
in ALFWorld, the same loss does not force additional reacquisition under this adapter.
This boundary is consistent with --- indeed predicted by --- our recoverability account:
compression cost is not a property of the operator, but of \emph{what execution-relevant state
becomes unavailable and how that state can be reacquired}.

\begin{figure}[t]
\centering
\includegraphics[width=0.95\textwidth]{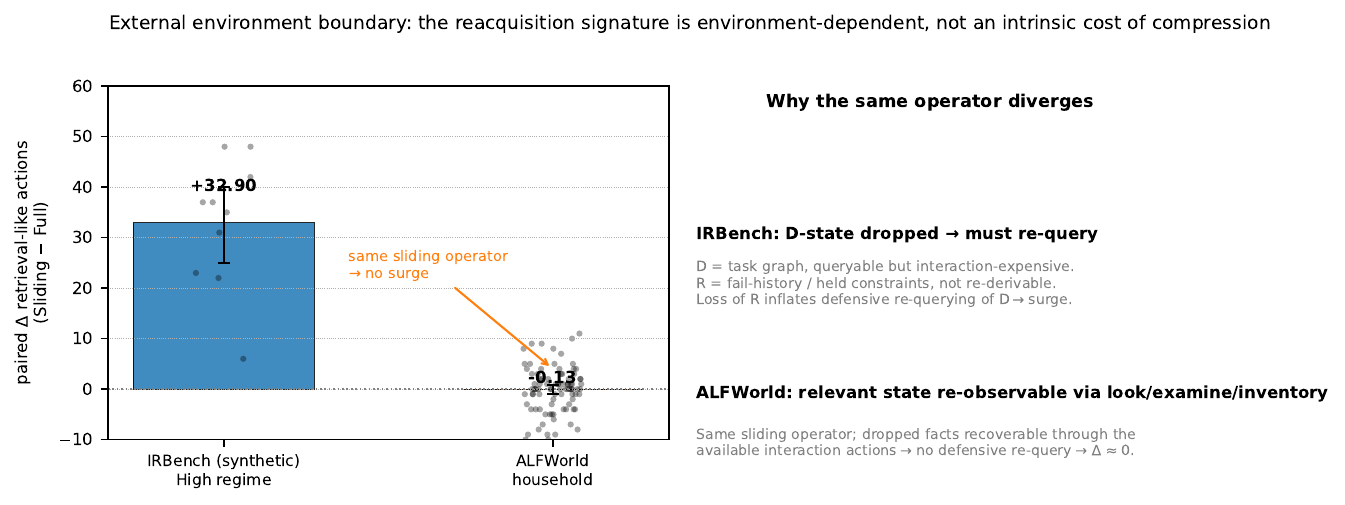}
\caption{External environment boundary: paired $\Delta$ retrieval-like actions under the same
sliding operator are large and positive in IRBench and $\approx 0$ (symmetric) in ALFWorld.}
\label{fig:crossenv}
\end{figure}

\subsection{The interaction budget: turn horizon and tool budget}
\label{sec:budget-horizon}

Our auxiliary tool budget ($\alpha \times$ reference, $\alpha = 2.0$) is a diagnostic, not a termination
criterion --- the agent is always cut off by the 24-turn horizon. The two quantities tell
different halves of the story (DeepSeek, High):

\begin{itemize}
\item \textbf{Turn horizon is binding in every High condition}, including Full: 7/10 Full runs and
  8--9/10 compressed runs terminate at \texttt{max\_turns}. Compression does not \emph{introduce} turn
  pressure; it reduces how much task progress the agent fits into the same horizon
  (completion 83\% $\to$ 66\% at $10\times$).
\item \textbf{The tool budget is the cost marker}: Full stays within the $\alpha \cdot 2.0$ budget in 10/10
  runs; Sliding $5\times$ and $10\times$ stay within it in 0/10 runs. The reacquisition overhead is
  large enough to violate a reference execution budget that the full-context agent never
  approaches.
\end{itemize}

Termination reasons are recorded per run (all\_complete / max\_turns), so failed runs can
be audited as turn-exhaustion rather than API errors; every reported condition has 10/10
completed runs.

\begin{figure}[t]
\centering
\includegraphics[width=0.95\textwidth]{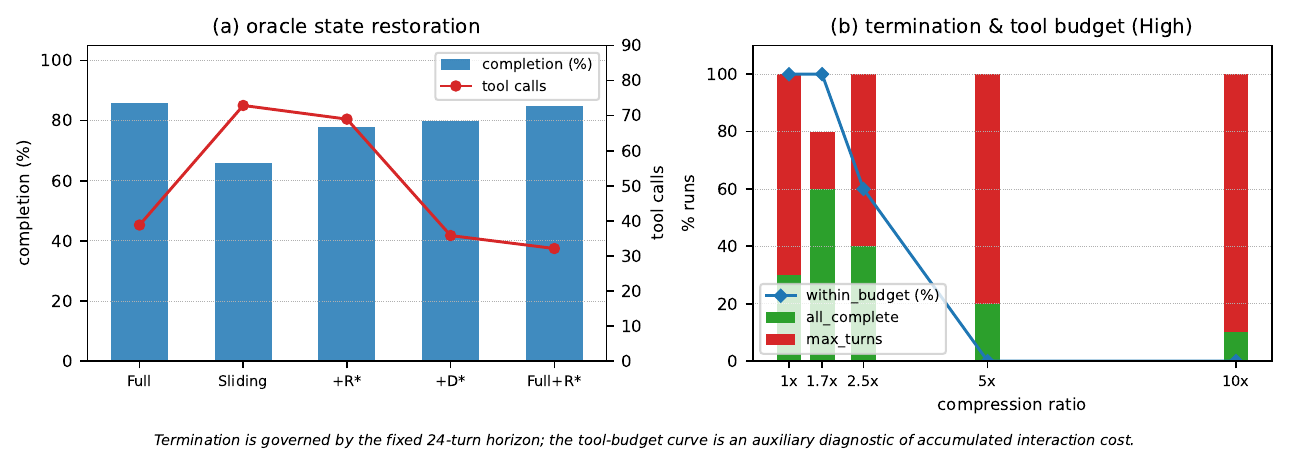}
\caption{Oracle conditions and the interaction budget: retrieval / execution decomposition,
termination (all\_complete vs.\ max\_turns), and within-budget diagnostics.}
\label{fig:oracle}
\end{figure}

\section{Discussion}
\label{sec:discussion}

\subsection{What the protocol measures, and why the metric matters}

Our protocol decomposes the interaction cost of compression into state reacquisition
(retrieval) and task work (execution). The headline observation is that \textbf{task completion
is an incomplete measure of this cost}: across our six cells, retrieval increases are
significant in five of six after Holm correction, while completion changes are not
significant in any cell. GPT-5.5 is the sharpest illustration --- completion statistically
unchanged ($p = 1.0$) while retrieval roughly triples. Completion is not useless: it
becomes a meaningful signal when reacquisition consumes enough of the interaction horizon
to limit task progress, as for DeepSeek under high compression. But used alone, it
under-states compression's cost on robust models and over-states it on fragile ones.
This is not an argument that compression ``hurts'' agents --- our own data rule that out ---
but that the cost is real, decomposable, and visible in tool behavior before (or without)
any change in completion. The mechanism behind the visible cost is a shift in \emph{where} the
agent spends its budget: under compression, the dominant shift is toward retrieval rather
than execution, and that reacquisition is what the completion signal can miss.

\subsection{Retention versus reacquisition: two separable sources of degradation}

The contrast between our Sliding and Summary operators isolates the lever: with the same
budget, the operator that preserves observed state facts is near-lossless, while the
operator that drops them degrades completion and triples retrieval. This separates
\emph{retention} (does the compressed artifact still contain the state?) from \emph{reacquisition}
(what does the agent do when it does not?). The two are complementary to the
attention-centric account of lost-in-compaction: that line shows surviving text can be
ignored; we show that \emph{absent} state must be re-fetched, at a budget cost that is
independent of attention. Runtime optimizations targeting one source of degradation will
not address the other.

Why is the reacquisition cost dominated by retrieval? The oracle intervention
(Sec.~\ref{sec:oracle}) isolates the mechanism. Restoring the externally queryable task graph ($D^*$) removes most
of the compensatory retrieval (Sliding 72.9 $\to$ 35.8 tool calls, $-51\%$; paired $p = .002$) and
recovers most of the completion gap (66\% $\to$ 80\%); restoring history-dependent state ($R^*$)
improves completion (+12pp) but leaves retrieval nearly unchanged (72.9 $\to$ 69.0, $-5\%$;
n.s.). The asymmetry follows from the environment's recoverability structure
(Sec.~\ref{sec:conditions}): D is queryable --- a query restores it --- so when it is lost the agent can recover it only by
re-querying the environment, an interaction pattern documented in production agent
systems as redundant retrieval or retrieval thrashing; R is history-dependent --- no single
public query restores it --- so when it is lost re-querying cannot help and the agent
proceeds under uncertainty, and the measured retrieval cost does not grow. The data are
consistent with this account: the added retrieval under compression is almost entirely
D-reacquisition ($D^*$ restores it to near the full-context level), while completion depends
on both state classes (both $D^*$ and $R^*$ restore most of the completion gap). We describe
the pattern as a re-query loop because restoring the queryable state removes the added
retrieval; we do not claim to observe the agent's internal decision process.

The retention interventions (Sec.~\ref{sec:retention-results}) refine \emph{where} in this loop a retention decision
bites. If the reacquisition cost is a re-query loop, the natural question is whether the
runtime can reduce it by retaining the \emph{right} state. Our answer is two-layered, and it
separates two notions that retention policies usually conflate. \emph{Within our D-state
pool}, which real atoms are retained has little measurable marginal effect: random
selection matches an offline hindsight oracle
($\Delta +0.10$ retrieval calls, CI $[-2.3, +2.3]$) within a small budget of the best possible
selection, because the externally queryable atoms are homogeneous in reacquisition cost.
\emph{Whether the retained content is valid and task-relevant} does matter: replacing the D
atoms with semantically irrelevant state raises retrieval by 57\% while completion stays
statistically unchanged. State content is thus behaviorally load-bearing --- the agent does
not treat the digest as an inert anchor --- but the granularity at which selection can
exploit it is coarse (valid versus not), not fine (which valid atom). The same
intervention shows the effect is budget-gated: it is visible only when the digest
dominates the compacted window, so recency is a competitive approximation to
recoverability-aware retention when budgets are tight.

Taken together, the four findings form a hierarchy of what drives interaction cost under
compression --- \emph{state absence} (strong: dropping state inflates reacquisition), \emph{valid
content} (strong: replacing real D content with semantically irrelevant content raises
cost 57\%), \emph{fine selection} (weak: choosing among valid atoms is bounded near zero), and
\emph{effect visibility} (budget-gated: the content effect appears only when the digest
dominates the window). A retention decision is therefore consequential at the level of
validity and presence, not at the level of fine-grained ranking, in this regime.

Two explanations for the selection null are compatible with our data. The D atoms are
homogeneous --- each is recoverable by a single public query at comparable cost --- so
retaining any real subset with equal coverage has equal reacquisition value. Alternatively
(or additionally), the agent might not exploit fine-grained differences within a short
horizon. The D-Irrelevant result favors the former: at $B=265$ the agent demonstrably
reacts to the digest's D content (it re-queries more when the content is fake), so it is
content-sensitive, and the within-class flatness is better attributed to D-atom
homogeneity than to an inability to use the digest at all.

\subsection{Implication for runtime design}

For a runtime designer, the actionable quantity is not ``which compression ratio is safe''
but \textbf{``which execution-relevant state, if dropped, would the agent spend its budget
re-acquiring?''} Our oracle results point to a concrete answer: externally queryable task
state (the task graph and its constraints) is a major source of reacquisition cost, and
restoring it removes roughly half the retrieval overhead. The retention interventions
qualify how to act on this: preserving history-dependent state (R) is strongly favored
when the retention budget is tight --- it is compact enough to retain fully at any budget,
and dropping it is what inflates defensive re-querying of D. Within the remaining D
budget, fine-grained ranking buys little in our setting; what matters is that the
retained content is real and task-relevant. Because the content effect is budget-gated,
recency is a competitive approximation to recoverability-aware retention under tight
budgets --- a boundary we state for this setting rather than a general law.

\textbf{A practical deployment checklist for diagnosing context-management cost.} The protocol
is not tied to our environment; it is a checklist any evaluator can apply to a
context-management strategy. In practice: (1) hold the agent, task instance, and
interaction horizon fixed; (2) record completion and interaction cost jointly; (3) where
tool semantics permit, decompose interaction cost into retrieval and execution; (4) where
causal attribution is needed, restore the specific dropped state as a control; and (5)
report the capability outcome and the interaction overhead together. The conclusion
connects directly to Proposition~1: a strategy that preserves completion but
substantially increases reacquisition cost should not be treated as equivalent to the
baseline solely on completion metrics.

\subsection{The boundary: model, regime, and environment dependence}

The reacquisition pattern is consistent across the three models in our study; its
conversion into reduced completion is not (Fig.~\ref{fig:phasediag}). We read this as a feature: the cost
of compression is a property of the \emph{agent-environment interaction}, while whether that
cost becomes failure is a property of the \emph{model's} interaction pattern and the \emph{task's}
recoverability. Evaluations of compression that report only completion may give a distorted view of cost, depending on how readily the model absorbs the extra
interaction. Our
two-axis result (Fig.~\ref{fig:phasediag}a) makes this explicit: interaction cost and task completion are
related but not interchangeable outcomes, so a cost-level diagnostic is needed to
separate the two.

The environment adds a third boundary axis (Sec.~\ref{sec:alfworld}): in ALFWorld, where the relevant state
can be re-observed directly through the available interaction actions, sliding compression
produces no retrieval surge at all --- the cost signal is not an intrinsic consequence of
shortening context. The \emph{presence} of the cost, not only its magnitude, is therefore
conditional on what execution-relevant state becomes unavailable and how that state can be
reacquired. A complete account of compression cost thus spans three boundary conditions ---
model, task regime, and environment recoverability --- and completion-only evaluation can
miss the cost on every one of them.

\subsection{Relation to the broader literature}

We connect to four lines: methods that report cost as a byproduct (we decompose it);
recoverability frameworks that treat it as a representation property (we measure it
behaviorally, under a budget); attention-dilution accounts (we isolate a distinct,
absent-state failure mode); and aggregate efficiency evaluation (we ask a finer question:
where the additional interaction goes, and whether completion exposes it). The
rate-distortion surveys' named gap --- no shared budget axis across layers --- is precisely
where our operator $\times$ regime $\times$ model matrix sits.

We therefore do not claim that context compression is intrinsically harmful or
beneficial; we characterize the interaction cost it introduces and the conditions under
which that cost becomes consequential.

\section{Limitations}
\label{sec:limitations}

\begin{enumerate}
\item \textbf{Two bounded environments, both synthetic-adjacent.} The core mechanism is measured
  in IRBench, a synthetic constraint-planning domain (two regimes); an ALFWorld probe
  (Sec.~\ref{sec:alfworld}) shows the reacquisition signature is environment-dependent --- it disappears where the
  relevant state can be re-observed directly. We make no claim that specific cost
  magnitudes transfer to open-ended tasks, and future work should extend the probe to
  partially-observable, long-horizon environments (e.g., WebArena, SWE-bench) where
  reacquisition is consequential. The value is the \emph{protocol} and the \emph{decomposition},
  which are environment-agnostic.

\item \textbf{Three models.} We evaluate three model families (DeepSeek, Qwen, GPT-5.5), which is
  not exhaustive. Cross-model comparisons are descriptive rather than paired, and we
  avoid interpreting absolute tool/token counts across providers (serving conditions are
  not fully comparable). Serving details are recorded in the reproducibility appendix.

\item \textbf{The 24-turn horizon is a design choice.} We evaluate under a bounded interaction
  budget because real agents always operate under one; an evaluation metric that is
  insensitive to a large interaction-cost increase \emph{within} that budget is a structural
  limitation of the metric, not merely a matter of ``not waiting long enough.'' A different
  horizon could shift where the cost--failure conversion falls; we report the conversion
  as regime-dependent rather than universal, and note that the D-Irrelevant intervention
  (Sec.~\ref{sec:retention-results}) changes cost sharply (and even raised completion directionally) within the same
  horizon --- the completion signal did not expose the cost difference. A horizon sweep is
  left to future work.

\item \textbf{Tool-call count is a proxy for ``cost''.} We deliberately measure the budgeted
  interaction currency --- turns and tool calls --- rather than provider-dependent wall-clock
  latency or dollar cost. This is a design choice: wall-clock and dollar costs are
  provider- and serving-dependent and would obscure the structural reacquisition signal
  we isolate.

\item \textbf{The retrieval/execution decomposition is tool-level.} It attributes each call to a
  schema, not to a semantic intention. The oracle results mitigate this (restoring state
  reduces the re-query volume), but a per-message intention analysis is out of scope.
  Turn-level temporal analysis of when retrieval bursts occur was likewise not feasible:
  per-run tool-call logs were aggregated to condition-level totals before storage.
  Instrumenting per-turn logging to study the temporal dynamics of reacquisition is left
  to future work.

\item \textbf{The extractive summary is our designed control, not a generic summarizer.} Its
  near-losslessness at $5\times$ is a property of the fact-preserving digest we define, and is
  the intended contrast to Sliding; it is not evidence that arbitrary LLM summarizers
  are lossless.

\item \textbf{Statistical power at 10 seeds.} The absence of significant completion changes at
  the pre-specified $5\times$ point is itself part of the finding: it shows completion is less
  sensitive to compression than the retrieval signal, which is significant in five of
  six cells after Holm correction. We treat the tool-level effect as the primary
  empirical result, report effect sizes and CIs for all comparisons, and note that the
  completion-side numbers are noisier. Pre-registered primary comparisons and Holm
  correction limit multiple-comparison risk.

\item \textbf{Single cross-model compression point.} The three-model matrix evaluates one
  compression ratio ($5\times$), fixed in the original experiment design and applied uniformly;
  only the DeepSeek condition has a full severity sweep (Sec.~\ref{sec:sweep}). This bounds how precisely
  we can compare completion across models at more aggressive ratios; the sweep is
  reported in full, including the $10\times$ point where DeepSeek completion first becomes
  significant.

\item \textbf{Two operators.} We evaluate a dropping operator (sliding window) and a
  fact-preserving extractive summary. The cost patterns of other operators ---
  LLM-generated abstractive summaries, token-level compressors --- remain to be
  characterized; we do not claim the specific magnitudes generalize.

\item \textbf{The D-Irrelevant fabrication is one kind of irrelevance.} The control
  (Sec.~\ref{sec:retention}) replaces D atoms with fabricated out-of-universe entities, which the agent can detect
  as non-existent. The added retrieval could therefore reflect verification of
  inconsistent state content as well as semantic irrelevance per se; both readings
  support the load-bearing claim, but the finer decomposition (e.g., an ``incorrect
  values on real entities'' control) is left to future work. We also do not claim the
  effect is specific to ``semantically irrelevant state'' as opposed to any corrupt
  content.

\item \textbf{The selection null and content replication are each confirmed on a subset of
  models.} The selection null (Random $\approx$ Hindsight) and the coverage dose-response are
  measured on DeepSeek only; the GPT-5.5 subset replicates the content intervention
  (Sec.~\ref{sec:crossmodel}) but does not re-run the selection controls, so we do not claim the selection
  null is cross-model. The content effect is confirmed on two models (DeepSeek, GPT-5.5)
  at the loose budget; its magnitude and budget-gating remain model-specific.
\end{enumerate}

\appendix
\section{Seed-level association}
\label{app:seed}

We examine whether, within a model--regime cell, the per-seed cost increase is associated
with the per-seed completion change, and whether this association varies across cells.
For each of the six model--regime cells we computed, per seed, the paired change in
retrieval calls ($\Delta C_R = C_R(\text{Sliding}) - C_R(\text{Full})$) and completion ($\Delta Q$, pp) at $5\times$, and the
Spearman correlation between them.

\begin{table}[h]
\centering
\small
\caption{Seed-level association between per-seed $\Delta C_R$ and $\Delta Q$.}
\label{tab:seed}
\begin{tabular}{@{}lrrrl@{}}
\toprule
Model & Regime & $\rho$ & p & 95\% bootstrap CI \\
\midrule
DeepSeek & High & $-0.40$ & .25 & $[-0.97, +0.49]$ \\
DeepSeek & Low & --- & --- & completion invariant across seeds \\
Qwen & High & $-0.32$ & .37 & $[-0.75, +0.29]$ \\
Qwen & Low & $+0.16$ & .67 & --- \\
GPT-5.5 & High & $-0.64$ & .045 & $[-0.84, -0.18]$ \\
GPT-5.5 & Low & --- & --- & completion invariant across seeds \\
\bottomrule
\end{tabular}
\end{table}

The direction is negative in all three High-regime cells, and GPT-5.5 High reaches
nominal significance ($\rho = -0.64$, $p = .045$) with a bootstrap CI excluding zero: seeds with
larger retrieval increases tend to show larger completion decreases, consistent with
reacquisition consuming the interaction budget. The two Low-regime cells are undefined
because completion is invariant across seeds (every seed completes all tasks in both
conditions). We report these as exploratory, descriptive within-seed associations, not
causal estimates; given the small per-cell sample ($n = 10$) and multiple comparisons, the
single significant cell should be read cautiously. We do not pool model--regime cells,
because cross-cell heterogeneity in baseline difficulty and model response makes a pooled
association difficult to interpret. The negative High-regime direction is consistent with
--- but does not establish --- the re-query loop account of Sec.~\ref{sec:discussion}.

\section*{Acknowledgments}
Portions of this manuscript, such as language translation and polishing, were developed
with the assistance of AI language models. All experimental design, implementation, data
collection, and analysis were conducted independently by the author.

\bibliographystyle{unsrt}

\end{document}